\documentclass[conference]{IEEEtran}
\IEEEoverridecommandlockouts
\usepackage{cite}
\usepackage{amsmath,amssymb,amsfonts}
\usepackage{algorithmic}
\usepackage{graphicx}
\usepackage{textcomp}
\usepackage{xcolor}
\def\BibTeX{{\rm B\kern-.05em{\sc i\kern-.025em b}\kern-.08em
    T\kern-.1667em\lower.7ex\hbox{E}\kern-.125emX}}
\begin{document}

\title{End-To-End Bias Mitigation: Removing Gender Bias in Deep Learning}

\author{\IEEEauthorblockN{Tal Feldman} \thanks{Both authors contributed equally to this work. The authors thank Dr. Sarra Alqahtani for her support and comments on the manuscript.}
\IEEEauthorblockA{\textit{Wake Forest University} \\
Winston Salem, USA \\
feldt19@wfu.edu}
\and
\IEEEauthorblockN{Ashley Peake}
\IEEEauthorblockA{\textit{Wake Forest University} \\
Winston Salem, USA \\
peakaa19@wfu.edu}
}

\maketitle

\begin{abstract}
Machine Learning models have been deployed across many different aspects of society, often in situations that affect social welfare. Although these models offer streamlined solutions to large problems, they may contain biases and treat groups or individuals unfairly based on protected attributes such as gender. In this paper, we introduce several examples of machine learning gender bias in practice followed by formalizations of fairness. We provide a survey of fairness research by detailing influential pre-processing, in-processing, and post-processing bias mitigation algorithms. We then propose an \textit{end-to-end bias mitigation} framework, which employs a fusion of pre-, in-, and post-processing methods to leverage the strengths of each individual technique. We test this method, along with the standard techniques we review, on a deep neural network to analyze bias mitigation in a deep learning setting. We find that our end-to-end bias mitigation framework outperforms the baselines with respect to several fairness metrics, suggesting its promise as a method for improving fairness. As society increasingly relies on artificial intelligence to help in decision-making, addressing gender biases present in deep learning models is imperative. To provide readers with the tools to assess the fairness of machine learning models and mitigate the biases present in them, we discuss multiple open source packages for fairness in AI.
\end{abstract}


\section{Introduction}

Research in artificial intelligence (AI) and machine learning (ML) has brought about significant advances in recent years. Beyond purely academic literature, ML models are increasingly being applied as decision-makers alongside (or even in lieu of) humans. Many of these models are used in applications with a social impact. For example, models have been deployed to determine optimal coinsurance rates and tax rates \cite{kasy_tax}, efficiently allocate health inspectors in cities \cite{glaeser}, decide which patients receive medical interventions based on predicted risk of heart failure \cite{bayati_data-driven_2014}, assess risk in numerous criminal justice decisions \cite{berk_criminal_2012}, and much more.

However, when ML models make such sensitive decisions, problems can arise. Specifically, predictive models may introduce or reinforce biases implicit in the data. When these models are indiscriminately deployed, they can have major impacts on individuals \cite{kashin}. As these models become more complex and the data more abundant, accuracy increases but transparency often decreases \cite{JANSSEN2016371}. Accordingly, questions of fairness in machine learning -- such as how to measure it and how to improve it -- become all the more important. 

The contributions of this review are threefold. First, to our knowledge, this review is the first to focus specifically on gender bias in ML. We detail several examples of ML models that are gender biased to motivate further research in this area, formalize notions of fairness in ML, and survey a number of algorithms for mitigating gender bias. These algorithms can be broadly categorized into three groups based on when during training they are executed: pre-processing, in-processing, and post-processing. Second, we present a novel fusion of these methods to create an \textit{end-to-end bias mitigation} framework for increasing fairness, which we test on a gender-biased dataset. Third, while the pre-, in-, and post-processing algorithms that comprise our end-to-end approach were applied to statistical models \cite{PSEs, Disparate_impact_remover, adversarial_debiasing}, we extend these algorithms to a deep learning setting to analyze the success of these techniques in complex domains. We close with a discussion of future directions of the field along with open-source packages that make fairness research broadly accessible.


\section{Gender Bias in Machine Learning}



When machine learning models are designed to optimize only one performance metric, such as profit or accuracy, they can have inadvertent but detrimental consequences \cite{balancing_competing_objectives}. When these models produce discriminatory results based on sensitive traits such as gender, we consider them to be 'biased' or 'unfair.'
Examples of gender-based unfairness in real applications are abundant. For instance, the Gender Shades project, a black-box algorithmic audit of three commercial facial analysis models using their Application Programming Interfaces (APIs), found that the classifiers performed better on male faces than female faces and lighter faces than darker faces. All of the classifiers in the study performed worst on Black, female faces. The Microsoft and IBM facial classifiers performed best on white, male faces \cite{gender_shades}. Several months after the study, the three audited companies updated their facial recognition APIs, decreasing accuracy disparities between lighter and darker skinned and male and female individuals \cite{actionable_gender}. The Gender Shades project thus illustrates the importance of algorithmic fairness research for changing the industry.

Machine translation has also been shown to be gender biased. For example, Prates et al. \cite{machine_translation} created simple sentences in 12 gender neutral languages of the form "He/she is a teacher" (where "teacher" is an occupation selected from a list of jobs) and then translated them into English using Google Translate. They found that Google Translate was biased against women when picking which gender pronoun to use, choosing the pronoun "he" over "she" for occupations in STEM, legal, corporate, and several other fields. Moreover, the study found that Google Translate amplified gender disparities, yielding male pronouns instead of female ones even more than reflected in the training data. After the paper's placement on arXiv.org and its increasing coverage in the media, Google began offering both masculine and feminine translations, once again indicating the importance of research on gender bias in machine learning.

In natural language processing (NLP), gender biases have been found in examples of word embedding, a method that vectorizes natural language to portray semantic similarities by vector proximity. Even embeddings trained rigorously on large datasets like Google News have been shown to be inherently sexist. For example, when Bolukbasi et al \cite{bolukbasi_man_is_to_computer_programmer} trained a model to solve analogy puzzles, the model returned that "man is to computer programmer as woman is to homemaker." Similarly stereotypical outputs of the model included relating man to "surgeon" and woman to "nurse," man to "shopkeeper" and woman to "housewife," and man to "brilliant" and woman to "lovely." 

In the realm of search engines, Datta et al. \cite{datta2015automated} found that Google displayed fewer high-paying jobs to women compared to men. In healthcare, AI algorithms have been deployed to predict personalized preventative and therapeutic care based on genetic and environmental factors. Many of these models discriminate against women by misdiagnosing them or ignoring relevant links between sex and health differences \cite{article}. This bias in model prediction can lead to sub-optimal (and sometimes even fatal) outcomes. 
Thus, across applications training models on biased data can reinforce gender discrimination, ultimately leading to detrimental societal results.

\section{Formalizing Fairness}

While it is clear that the aforementioned examples constitute unfair machine learning practices, in order for research to mitigate unfair learning, it is necessary to formally define what is fair and what is not.

In general, fair models should indicate the absence of discrimination with respect to \textit{protected classes}. For instance, in the United States, a number of classes or attributes have been defined as "protected" by law. Protected attributes, according to the Fair Housing and Equal Credit Opportunity Acts \cite{fair_housing_act_1968, equal_credit_act}, include gender identity, age, color, disability, familial status, marital status, national origin, race, religion, sex, and public assistance status. For the purpose of this review, we specifically focus on metrics to evaluate fairness based on sex.\footnote{For notional simplicity and following the standard of other fairness work, we consider sex as a binary variable with values "male" and "female." However, we note that gender and sex are not strictly binary and the same ideas of fairness presented here can be extended accordingly.}


While there are a multitude of fairness definitions in the literature, this paper considers the most widely used definitions, motivated by \cite{verma_definitions_explained} and \cite{Gajane2017OnFF}. The broadest definition of fairness is simply "the absence of any prejudice or favoritism towards an individual or a group based on their intrinsic or acquired traits" \cite{ninareh}. Specific fairness definitions can be separated into two main categories: individual fairness and group fairness. 

\subsection{Individual fairness}
For a model to be fair at the individual level, similar individuals must receive similar predictions. From a social science perspective, this idea is closest to the notion of fairness as consistency \cite{binns2019apparent}. A simple formalization of individual fairness is fairness through unawareness.

\vspace{5pt}
\noindent \textbf{Definition 1: Fairness through unawareness  \cite{kusner2018counterfactual}}

\noindent\textit{A model $\hat{Y}$ achieves} \textit{fairness through unawareness} \textit{if the protected attribute $A$ (i.e. sex) is not utilized to make predictions.}

\vspace{5pt}
\noindent In this case, the model is "unaware" of the sex of an individual, so we expect that similar individuals with different sexes should still receive similar predictions. 
A more generalized definition of individual fairness is counterfactual fairness, which is achieved if the prediction does not change if an individual's sex is changed to its counterfactual \cite{kusner2018counterfactual}.

\vspace{5pt}
\noindent \textbf{Definition 2: Counterfactual fairness \cite{kusner2018counterfactual}}

\noindent\textit{Formally, a predictor $\hat{Y}$ is} \textit{counterfactually fair} \textit{if for any attributes $X=x$ and sex $A=a$,}
\begin{equation*}
    \begin{aligned}
      P(\hat{Y}_{A \leftarrow a} = y \;|\; X=x, A=a)
    \\= P(\hat{Y}_{A \leftarrow a'} = y\; | \;X=x, A=a) 
    \end{aligned}
\end{equation*}
In other words, if we can change only the sex of an individual and still get the same model prediction, we have achieved counterfactual fairness. 

\subsection{Group fairness} 
Group fairness ensures that different groups receive the same predictions with close to equal probability. This is related to the philosophical idea of egalitarianism, particularly from a collectivist standpoint \cite{Gajane2017OnFF, binns2019apparent}. A practical implementation of group fairness is affirmative action policies. Statistical parity (sometimes called demographic parity) is a common formalization of group fairness \cite{kusner2018counterfactual}. This notion can be measured by the statistical parity difference.

\vspace{5pt}
\noindent \textbf{Definition 3: Statistical Parity Difference \cite{kusner2018counterfactual}}

\noindent\textit{The statistical parity difference of a model is}
\begin{equation*}
    \begin{aligned}
    SPD = P(\hat{Y} = 1\;|\;A=\text{male}) - \\P(\hat{Y} = 1\;|\;A=\text{female})
    \end{aligned}
\end{equation*}
\textit{The model achieves perfect fairness if} \textit{$SPD =0$.}
\vspace{5pt}

\noindent We can similarly measure a model's group fairness using the equal opportunity difference, which is simply the difference in true positive rates for protected and unprotected groups \cite{Gajane2017OnFF}. 

\vspace{5pt}
\noindent\textbf{Definition 4: Equal Opportunity Difference \cite{hardt2016equality}}

\noindent\textit{Formally, for a predictor $\hat{Y}$ with true label $Y$,} \textit{the equal opportunity difference is }
\begin{equation*}
    \begin{aligned}
    EOD &= P(\hat{Y} = 1\;|\;A=\text{male}, Y=1) 
    \\&- P(\hat{Y} = 1\;|\;A=\text{female}, Y=1)
    \end{aligned}
\end{equation*}
\textit{The model achieves perfect fairness if} \textit{$EOD =0$.}

\vspace{5pt}
\noindent This measurement indicates how well the model performs on the entire group of males versus females. Another, similar metric of group fairness is the average odds difference, which additionally incorporates false positive rate.

\vspace{5pt}
\noindent\textbf{Definition 5: Average Odds Difference \cite{aif360-oct-2018}}

\noindent \textit{The average odds difference} \textit{of a model $\hat{Y}$ is}
\begin{equation*}
    \begin{aligned}
    AOD =\;&\frac{1}{2}\;[(FPR_{A=\text{female}} - FPR_{A=\text{male}}) 
    \\&+ (TPR_{A=\text{female}} - TPR_{A=\text{male}})]
    \end{aligned}
\end{equation*}
\textit{where $FPR$ is the false positive rate and $TPR$ is the true positive rate. The model achieves perfect fairness if $AOD=0$}

\vspace{5pt}
\noindent In other words, this is a measure of how different the true versus false positive rates are for each group. Another widely used group fairness metric is disparate impact \cite{Disparate_impact_remover}.

\vspace{5pt}
\noindent\textbf{Definition 6: Disparate Impact \cite{Disparate_impact_remover}}

\noindent \textit{The disparate impact} \textit{of a model $\hat{Y}$ is}
\begin{equation*}
    \begin{aligned}
    DI =\frac{P(\hat{Y} = 1 | A = \text{female})}
    {P(\hat{Y} = 1 | A = \text{male})}
    \end{aligned}
\end{equation*}
\textit{The model achieves perfect fairness if $DI = 1$.}
\vspace{5pt}

\noindent Notice that this metric measures the same relationship as statistical parity difference, just formulated as a proportion. We include it here in addition to statistical parity difference as a commonly reported way to quantify the same effect.


\section{Method}
$\label{gen_inst}$

Although the problems of gender-focused algorithmic bias outlined above are complex, mitigating these problems is not an insurmountable challenge. Indeed, Dr. Jennifer Chayes, the former Managing Director of Microsoft Research writes that "with careful algorithm design, computers can be fairer than typical human decision-makers, despite the biased training data" \cite{huffpost}. It is with this ideal in mind that we now explore ways to allay the unfairness discussed above. In this section, we first introduce benchmark datasets for evaluating bias mitigation algorithms and then discuss several of these algorithms to provide an overview of current work in the field.  

\subsection{Benchmark Datasets}

One of the most widely used benchmark datasets for studying gender fairness in machine learning is the Adult dataset, extracted from the 1994 U.S. Census database and available from the UCI Machine Learning Repository \cite{Dua:2019}. Many researchers test their bias mitigation techniques on the Adult dataset, including \cite{Disparate_impact_remover, mehrabi2019survey, besse2020survey, kamishima_fairness-aware_2012, nabi2018fair, DBLP:journals/corr/ZhangWW16a, NIPS2017_9a49a25d, zhang, Zafar, wu2018fairnessaware, post_eq_calib}. Models trained on this dataset often aim to predict a binary variable representing whether a person's income is above or below \$50,000 per year. The dataset contains $48,842$ observations and a variety of features representing demographic information including sex, ethnic origin, age, and education level. There are two sensitive features in this dataset on which models are often biased: sex and ethnic origin \cite{besse2020survey}. For our purposes, we focus on sex as the sensitive feature. Accordingly, in Section 5 we run experiments on the UCI Adult dataset to evaluate the gender fairness of several models, omitting the ethnic origin feature entirely as in \cite{Disparate_impact_remover}.

Another benchmark dataset worth mentioning is COMPAS, which is used to build predictive models for the likelihood that a defendant will recommit a violent crime. It contains over 100 features, including race, gender, and criminal history. The COMPAS dataset was compiled to assist judges in making many criminal justice decisions but was found to be incredibly unfair. Models trained on COMPAS dataset without any considerations of fairness were far likelier to classify Black defendants as high risk than white ones \cite{compas_paper, propublica_story}. Although COMPAS deals with race discrimination, bias mitigation algorithms tested on COMPAS can be extended to reducing gender-based unfairness. Thus, several models reviewed here were originally tested on the COMPAS dataset \cite{PSEs,nabi2018fair, NIPS2017_9a49a25d, post_eq_calib, compas_paper, DBLP:journals/corr/abs-1807-00199, pmlr-v80-komiyama18a}.


\subsection{Bias Mitigation Algorithms}

Bias mitigation algorithms can be broadly split into three main approaches: algorithms that reduce bias before model training (pre-processing), during model training (in-processing), or after model training (post-processing). 
To provide an overview of the current state of the field, we discuss representative examples of these three methods, specifically focusing on domains in which models have high propensity for gender discrimination. Most of the techniques described were presented only on statistical learning models. We extend this work by applying the algorithms to our own deep learning model in Section 5.

\subsection{Pre-processing: bias mitigation through data manipulation}

Biases in ML models are often reflections or amplifications of biases already inherent in the training data. Motivated by this fact, researchers have attempted to increase model fairness by directly altering the data distribution \cite{PSEs,nabi2018fair,   DBLP:journals/corr/ZhangWW16a, NIPS2017_9a49a25d,DBLP:journals/corr/abs-1903-08136, pre-process}. These methods first quantify the discriminatory effects within the data to subsequently remove or account for them. The specific mechanism for handling discrimination in the data differs across applications, but each aims to create a fair training distribution. 

One of the primary benefits of pre-processing is that it is done independently of the model itself, and thus can be used in a black box setting. Furthermore, by changing the data before ever building a model, this technique addresses the root of the fairness issue. However, in some applications it requires unrealistic assumptions about the training distribution or results in the loss of too much information implicit in the original data. To further analyze this technique, we detail two representative pre-processing methods.

In \cite{nabi2018fair}, Nabi and Shpitser consider fairness in probabilistic classification and regression as a constrained optimization problem. They employ causal inference in order to quantify the effects of a variable $A$ on an outcome $Y$ -- both directly ($A \xrightarrow[]{} Y$) and indirectly through causal pathways ($A \xrightarrow[]{} M \rightarrow W \xrightarrow[]{} Y$). These indirect pathways, designated \textit{path-specific effects} (PSEs), form the basis of their approach. Specifically, they formalize discrimination as the presence of certain PSEs. They can then bound the PSE, effectively transforming the inference problem on a distribution $p(Y, X)$ into an inference problem on another distribution $p^*(Y, X)$. Thus, $p^*$ serves as the hypothetical mirror of the distribution $p$ in a "fair world." They approximate $p^*$ by solving a constrained maximum likelihood problem:


\begin{equation}
\begin{aligned}
    \hat{\alpha} = \arg\max_{\alpha} \mathcal{L}_{Y,\mathbf{X}}(\mathcal{D}; \alpha)\\
    \text{subject to } \epsilon_{l} \leq g(\mathcal{D}) \leq \epsilon_{u}
    \label{eq:constrained-optimization}
    \end{aligned}
\end{equation}

\noindent where $\mathcal{D}$ is a finite set of samples from the original distribution $p$, $\mathcal{L}_{Y,\mathbf{X}}(\mathcal{D}; \alpha)$ is a likelihood function parameterized by $\alpha$, $g(\mathcal{D})$ is an estimator of the relevant PSE, and $\epsilon_{l}, \epsilon_{u}$ are constraints on the PSE. 
  
A model can then trained on this fair distribution $p^*$, effectively reducing discrimination. To test this method, they train a Bayesian Additive Regression Tree (BART) model on the COMPAS dataset for prediction of recidivism and on the Adult dataset for prediction of income level. Discriminatory PSEs are domain specific and were pre-determined by experts in the specific application. In both of these cases, their constrained optimization approach removed discrimination but did reduce inference accuracy. Specifically, in the unconstrained model trained on the Adult dataset, the original PSE was 3.16, meaning that if a female were instead male, \textit{ceteris paribus}, she would be 3 times more likely to have a higher income. The constrained model limits this PSE to the fair range (0.95, 1.05) and reduces model accuracy from 82\% to 72\%. They do not test this approach on any deep learning models, leaving parameterization in nonlinear settings to future work. Nonetheless, their results prove the feasibility of parameter constraints for learning fair classifiers.

In \cite{Disparate_impact_remover}, Feldman et al. propose another technique to increase fairness by altering the initial data distribution. They consider the conditional distributions $F_x(y)$ for each protected class (i.e. sex) separately and define a distribution $F_A(y)$ to be the median of the these. A fully repaired distribution is created such that for each $y\in Y_x$, the repaired $\bar{y}$ falls at the same percentile of the median distribution as $y$ did in its conditional distribution based on sex. In this way, the fair distribution preserves relative ordering. Because this fully repaired distribution can degrade predictive accuracy, they also propose a partial repair method. Specifically they define a Geometric Repair:
\begin{equation}
    \bar{F_x}^{-1}(\alpha) = (1-\lambda)F_x^{-1}(\alpha) + \lambda(F_A)^{-1}(\alpha)
\end{equation}
where $\lambda$ is the level of repair. In this way, they compute a linear interpolation in the original data space to produce partially repaired values that are near their respective fully repaired values. 

To test this method, they train a logistic regression, support vector machine (SVM), and Gaussian naive Bayes (GNB) on the Adult dataset. Using their geometric repair algorithm, they are able to reach a fair disparate impact value ($DI \geq .8$) for each model and a BER (indicating lack of predictability) of approximately 0.45 compared to 0.38 for the unmitigated baseline.

\subsection{In-Processing: Bias Mitigation Through Model Training}

To account for the limitations of pre-processing, some researchers have instead proposed fair training algorithms as a method to reduce bias in various machine learning models. These are, in effect, "online" ways to improve fairness, producing nondiscriminatory results from biased data. Fair model training is primarily accomplished in one of two ways: updating the objective function or imposing constraints on the model. We illustrate both of these approaches, first focusing on adversarial debiasing as an effective example of manipulating the objective function.


Adversarial learning \cite{lowd} has been proposed as a mechanism for reducing bias in many different applications \cite{zhang,DBLP:journals/corr/abs-1807-00199, Sweeney, FairnessGAN}. In adversarial debiasing, an adversarial network is trained to predict protected demographic information from biased labels. Adversarial learning is then harnessed in order that the fair model learns to decorrelate the protected data from potential biases. This idea originates from the work of Goodfellow et al. \cite{goodfellow2014generative}, who proposed a Generative Adversarial Network (GAN) framework. Although formulated for a very different application (specifically, generating images), Goodfellow's technique of using multiple competing networks in order to train one model to deceive another can be extended to our problem domain. Under this framework, the adversarial network acts as a \textit{discriminator} in a typical GAN. The fair network must then learn to fool the discriminator -- that is, to reduce the likelihood that the discriminator correctly predicts the protected attribute from the model's output -- while still maintaining its own accuracy. In this way, adversarial learning serves to reduce the impact that a protected trait has on the model's output, thus mitigating implicit biases in model predictions on account of that trait. 

A benefit of this technique is that it is generalizable across datasets and applications. It can offer increased accuracy because it maintains the integrity of the data. Furthermore, it requires no assumptions about the distribution of the dataset. However, it does require access to model parameters, making it impossible in a black-box setting. To further illustrate this adversarial debiasing method, we highlight a specific study that represents the current progress and direction of the field. 

Zhang et al. propose an adversarial learning framework for a classifier trained on the Adult dataset \cite{zhang}. They consider both the case in which gender, the protected attribute, is explicitly included in the data and the case in which gender is inferred from latent semantics, such as word embeddings. They implement a model that uses information $X$ to predict $Y$, and thus is trained by modifying weights $W$ to minimize the loss $L_p$, representing the difference between $y$ and $\hat{y}$. They then define an adversary that attempts to predict a protected attribute $Z$ from $\hat{Y}$. The adversary's weights $U$ are updated according to the loss $L_a$, representing the difference between $z$ and $\hat{z}$. We introduce all of this notation to clearly illustrate the update rule for $W$ according to the adversarial training method: 
\begin{equation}
    \nabla WL_p + proj_{\nabla WL_a}\nabla WL_p - \alpha \nabla WL_a   
    \label{eq:adversarial}
\end{equation}
where $\alpha$ is a tunable parameter to balance the trade-off between fairness and model accuracy. Intuitively, the first term helps reduce the loss of the predictive model and the third term helps increase the loss of the adversary. The middle term is included in order to prevent the model from moving in a direction that would actually aid the adversary. 

Using this adversarial debiasing method, the proposed logistic regression model in \cite{zhang} for predicting income achieved near \textit{equality of odds}, meaning that model predictions did not give any additional information about gender to the adversary. More formally, $\hat{Y}$ and $Z$ were independent given $Y$. Furthermore, the model's predictive accuracy reduced by only 1.5\%. Overall, Zhang et al. prove the efficacy of the adversarial debiasing method on a classification task.

Other researchers have suggested that the fairness problem can be solved during training by subjecting model parameters to a fairness constraint \cite{Zafar, wu2018fairnessaware, pmlr-v80-komiyama18a, gencoglu,  DBLP:journals/corr/ZhaoWYOC17}. In this method, a model is trained under a constrained optimization method to reduce bias reflected in the model's output distribution. We detail a specific approach to this idea in order to provide an overview of its possibilities for mitigating gender bias.

Zhao et al. propose a method for training a fair classifier by injecting corpus-level constraints \cite{DBLP:journals/corr/ZhaoWYOC17}. They focus on models for the NLP tasks of multi-label object classification and visual semantic role labeling, which tend to perpetuate and amplify biases implicit in the data. They build a Reducing Bias Amplification (RBA) framework that constrains the frequency of potentially-biased pairs (e.g. "woman" with "cooking") in model output to at most the rate of co-occurrence in the training distribution. The inference problem is thus formulated as a constrained inference problem:
\begin{equation}
\begin{aligned}
    \max_{\{y^i\}\in\{Y^i\}}& \sum_{i}f_{\theta}(y^i, i)\\
     \text{subject to } A&\sum_{i}y^i-b \leq 0.
\end{aligned}
\end{equation}
where $f_{\theta}(y, i)$ is a scoring function based on the trained model $\theta$, $A$ is a matrix of the coefficients of one constraint, $b$ is the desired gender ratio, and $\{Y^i\}$ is the space spanned by possible combinations of labels for all instances.
This method is notably distinct from other training-based approaches in that it does not consider the influence of individual training data on model output. Instead, it calibrates model predictions such that the classification outputs adhere to a fair distribution across the \textit{entire} test corpus. However, we still thematically group it with this section as it relies on the same principle of mathematically manipulating the model parameters to reduce correlations based on biased attributes. Overall, Zhao et al. trained a conditional random field model to show that this method effectively reduces bias amplification in these NLP tasks by over 40\%, without reducing the model's predictive accuracy more than 1\%.

\subsection{Post-Processing: Bias Mitigation Through Prediction Constraints}

Other researchers have proposed post-processing as a method for bias mitigation \cite{hardt2016equality, post_eq_calib, woodworth2017learning, decision_theory, lohia2018bias}. This method generally works by manipulating model predictions based on a fairness constraint. The primary benefit of this method is, like pre-processing, it does not require access to model parameters, so it can be applied in a black-box setting. This also means it can theoretically be utilized for any kind of machine learning model to increase run-time fairness. Some approaches to post-processing do not even necessitate access to the input features and can be applied to the joint distribution over labels $Y$ and model predictions $\hat{Y}$. However, this technique can result in a significant loss in performance and has been shown to be sub-optimal \cite{woodworth2017learning}. Furthermore, it increases fairness strictly with respect to the specified constraint -- not necessarily to any other notions of fairness \cite{post_eq_calib}. We further explore this technique by detailing an example approach.

Pleiss et al. propose Calibrated Equalized Odds \cite{post_eq_calib} as a post-processing algorithm. This technique attempts to retain a model's calibrated probability estimates while minimizing error disparities across different groups. In this method, a cost function is defined to penalize a model for disparities in false-negative rates, false-positive rates, or a weighted combination of these across different groups such as women and men. The choice of cost function condition is problem-specific and up to practitioners. For example, in experimenting with the Adult Dataset, \cite{post_eq_calib} uses the false-negative condition. A model achieves Calibrated Equalized Odds if the cost of calibrated predictions is the same for each group. They equalize this cost by suppressing predictive information for random subsets of groups with unequal classifier costs. This new, post-processed classifier is derived using a validation (or holdout) set and attains parity, but decreases accuracy by approximately 10\%.

\section{Experiments}
To analyze the bias mitigation algorithms discussed above, we present results from several experiments. We train a model to classify income level from the Adult dataset (Section 4.1) and compare examples of the pre-processing, in-processing, and post-processing algorithms on this model. Furthermore, we test combinations of these algorithms as a novel approach for increasing model fairness. 

\subsection{Model}
As fairness is still a growing field, many bias mitigation techniques are tested on only simple statistical models, rather than more complex ML models. However, neural networks and other complex ML techniques have become popular in recent years because of their capacity to model complicated phenomena \cite{neural_popular}. To evaluate the success of standard bias mitigation techniques for deep learning, we perform all of our tests using a neural network model. 

We train a network with 3 fully connected layers and 200 hidden units. We use the relu activation function, the Adam optimizer, and sigmoid cross-entropy loss function. We split the dataset into 70\% training, 15\% testing, and 15\% validation. Following \cite{aif360-oct-2018} and \cite{Disparate_impact_remover}, we use the features age, education, capital gain, capital loss, and hours-per-week. We train for 50 epochs with a batch size of 128 and learning rate of 0.001. Hyperparameters remain the same for each bias mitigation algorithm tested.

\subsection{Bias Mitigation}
To reduce the gender bias of our baseline model, we first implement bias mitigation algorithms discussed in Section 4. To compare the different techniques, we use the Disparate Impact Remover (pre-processing) \cite{Disparate_impact_remover}, Adversarial Debiasing  (in-processing) \cite{zhang}, and Calibrated Equalized Odds  (post-processing) \cite{post_eq_calib}. We use the false-negative rate cost function for Calibrated Equalized Odds, following \cite{post_eq_calib} in their experiments with the Adult Dataset. 

\begin{figure}[ht]
    \centering
    \includegraphics[width=7cm]{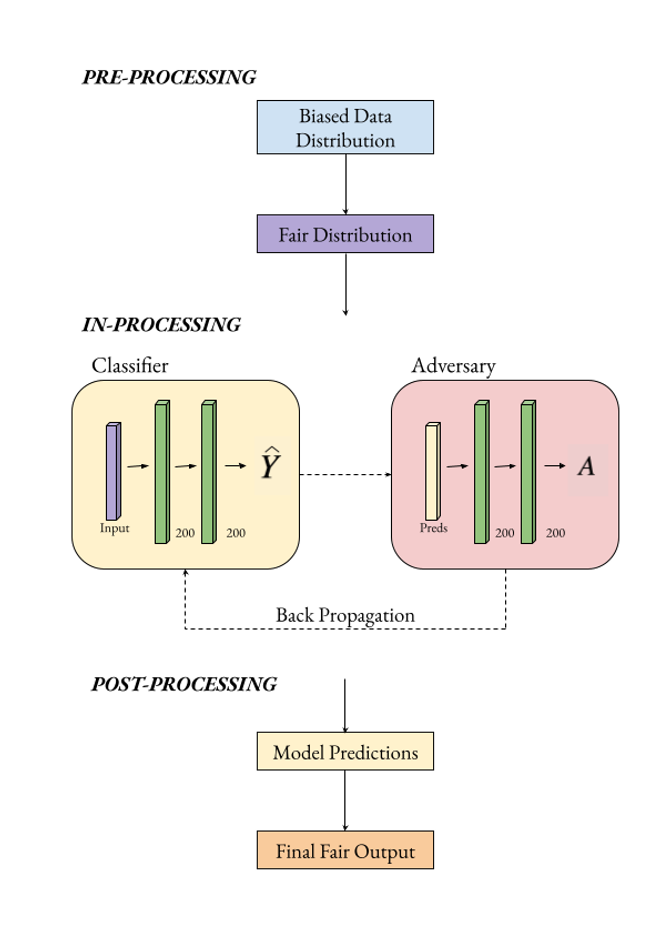}
    \caption{End-to-End Bias Mitigation Framework}
    \label{fig:end-to-end}
\end{figure}

To further improve fairness, we propose the fusion of multiple of these algorithms to create \textit{end-to-end bias mitigation} for deep learning. That is, we run each of pre, in, and post processing on the same model. We diagram this approach in Fig. \ref{fig:end-to-end}. We compare this method to the single method baselines described above.

\begin{table*}[t]

    \centering
        \caption{Average Metrics for Each Model Combination Over 100 Runs}

    \begin{tabular}{|c||c|c|c|c|}
    \hline
    Debiasing  & Classification & Statistical Parity & Equal Opportunity  & Average Odds\\
    Technique & Accuracy & Difference & Difference & Difference\\
    \hline
       \textbf{None} & \textbf{0.8317}&	-0.1943&-0.2881&-0.1872\\
        \hline
        \textbf{Pre} & 0.8310&-0.2057	&-0.3040	&-0.1997\\
        \hline
        \textbf{In} & 0.8124&	-0.0220	&0.1044	&0.0673\\
        \hline
        \textbf{Post} & 0.7784&	\textbf{-0.0062}&	0.1525&	0.0720\\
        \hline
        \textbf{Pre + In} & 0.8134&	-0.0294&	0.0894&	0.0577\\
        \hline
        \textbf{Pre + Post} &0.7778&	-0.0064&	0.1546&	0.0727\\
        \hline
        \textbf{In + Post} & 0.8109&	-0.0267	&0.0790&	0.0523\\
        \hline
        \textbf{Pre + In + Post} & 0.8113&	-0.0301&\textbf{0.0785}&	\textbf{0.0510}\\
        \hline
    \end{tabular}
    \label{tab:averages}
\end{table*}

\subsection{Results and Discussion}
The results for each algorithmic combination are presented in Table \ref{tab:averages}, where 'Pre,' 'In,' and 'Post' represent Disparate Impact Remover, Adversarial Debiasing, and Calibrated Equalized Odds, respectively. Since different metrics of fairness are used for different objectives \cite{post_eq_calib}, we report the Statistical Parity Difference, Equal Opportunity Difference, and Average Odds Difference. Note that models are considered fair on each of these metrics if they have a value in the range $[-0.1, 0.1]$.

From the results in Table \ref{tab:averages}, it is clear that there is no model that performs best on all fairness metrics. Intuitively, the unmitigated model has the highest accuracy but is unfair on all metrics. Interestingly, the 'Pre' model is actually less fair and less accurate than the unmitigated baseline. This is likely because the disparate impact remover utilizes a linear interpolation method, which is known to lead to a reduction in the training objective in neural network models \cite{lucas2021analyzing}. The 'In' model is fairer on every metric and is only around two percent less accurate compared to the baseline. As noted above, calibrated equalized odds is a sub-optimal algorithm, which is reflected in the 'Post' model's low accuracy. Still, 'Post' achieves the lowest Statistical Parity Difference. The poor performance of 'Post' in terms of Equalized Odds Difference serves as a reminder that these metrics are not equivalent and that practitioners should put ample consideration into which metric they optimize. 

Models trained on our fusion algorithms generally outperform the individual ones in terms of fairness, while retaining similar accuracy. All of the fusion algorithms except for 'Pre + Post' have an accuracy above 0.81 and are fair on all metrics. Furthermore, the end-to-end mitigated model ('Pre + In + Post') performs well across all test metrics, achieving the best results on Equalized Odds Difference and Average Odds Difference. Although the end-to-end mitigated model utilizes 'Post', with its low accuracy score, it achieves an accuracy of 0.81, which is comparable to the other models that substantially increased fairness.

Notably, while none of the individual algorithms ('Pre', 'In', and 'Post') achieve fairness on all three metrics in a deep learning setting, our end-to-end mitigated model does -- as did two other fusion algorithms. Thus the results indicate that by combining different algorithms we are able to leverage the strengths of each one, creating new algorithms that are fairer but similarly accurate. This suggests that combining bias mitigation algorithms to create end-to-end mitigated models is a viable approach to increasing model fairness, particularly for deep learning models.

\section{Recommendations}

Although questions of fairness are receiving more attention in the literature, there are numerous topics surrounding machine learning bias that warrant further research -- especially concerning gender. 
This paper suggests that a worthwhile direction for future work is studying complex models such as machine translation or deep reinforcement learning which would extend to broad, social impact applications beyond the benchmark datasets. Furthermore, as illustrated by the adversarial debiasing approach, many parallels can be drawn from the fairness question to more developed problem domains like privacy and security. These connections may be a highly promising source of advanced bias mitigation algorithms. Finally, we encourage further exploration of the end-to-end bias mitigation method presented in this paper as a promising framework for both academics and practitioners.


There are a plethora of open source fairness in AI tools and packages for those entering the field. We include a discussion of major libraries and tools for fairness in hope that it will provide a foundation for those doing applied work or research in any machine learning domain.

AI Fairness 360 (AIF360), developed by IBM Research and available in Python and R, is an off-the-shelf library that has a number of metrics on which to test models for biases and algorithms to mitigate these biases, including the ones in Table \ref{tab:averages}. Microsoft's Fairlearn assesses machine learning models for fairness, visualizes these metrics in an interactive dashboard, and contains numerous bias mitigation algorithms \cite{bird2020fairlearn}. Another tool that offers visual fairness analytics is \textit{FairSight}, which also includes bias measurement and mitigation features \cite{ahn2019fairsight}. Fairness Indicators, a package created by Google for assessing Tensorflow models, can compute standard fairness metrics for classifiers. Another popular package is Aequitas, which conducts bias audits on a multitude of metrics \cite{2018aequitas}. This large number of open source tools for fairness should make fairness applications and research more approachable for those entering the field.

\section{Conclusion}
In the eternal words of the late Justice Ruth Bader Ginsburg, "Women belong in all places where decisions are being made. It shouldn't be that women are the exception" \cite{rbg}. As artificial intelligence permeates society, it is incredibly important to remember that while these tools can be revolutionary, they have pitfalls. Without intentionally considering fairness, models may be biased on certain protected attributes -- such as race, religion, and sex. Examples of such biases in machine learning models that were actually deployed are plentiful. Thus, we urge researchers and practitioners to consider the fairness of their models, for computer science is still a field that severely lacks diversity, especially gender diversity \cite{diversity_pedagogy}.

To assist in this task of making machine learning models more gender equitable, we reviewed several bias mitigation algorithms and implemented them on a deep learning model, testing on the Adult dataset. We further introduced our \textit{end-to-end bias mitigation} approach, combining pre, in, and post-processing algorithms. We find that this end-to-end approach performed best on two fairness metrics without losing much accuracy. We hope that these exploratory results encourage others to investigate the implications of this study in applications beyond the Adult dataset and with other baseline models.
\bibliographystyle{ieeetr}

\bibliography{IEEEexample}

\begin{thebibliography}{10}

\bibitem{kasy_tax}
M.~Kasy, ``Optimal taxation and insurance using machine learning — sufficient
  statistics and beyond,'' {\em Journal of Public Economics}, vol.~167,
  pp.~205--219, 2018.

\bibitem{glaeser}
E.~L. Glaeser, A.~Hillis, S.~D. Kominers, and M.~Luca, ``Crowdsourcing city
  government: Using tournaments to improve inspection accuracy,'' {\em The
  American Economic Review}, vol.~106, no.~5, pp.~114--118, 2016.

\bibitem{bayati_data-driven_2014}
M.~Bayati, M.~Braverman, M.~Gillam, K.~M. Mack, G.~Ruiz, M.~S. Smith, and
  E.~Horvitz, ``Data-{Driven} {Decisions} for {Reducing} {Readmissions} for
  {Heart} {Failure}: {General} {Methodology} and {Case} {Study},'' {\em PLOS
  ONE}, vol.~9, p.~e109264, Aug. 2014.
\newblock Publisher: Public Library of Science.

\bibitem{berk_criminal_2012}
R.~Berk, {\em Criminal {Justice} {Forecasts} of {Risk}: {A} {Machine}
  {Learning} {Approach}}.
\newblock Springer Science \& Business Media, Apr. 2012.
\newblock Google-Books-ID: Jrlb6Or8YisC.

\bibitem{kashin}
K.~Kashin, G.~King, and S.~Soneji, ``Systematic bias and nontransparency in us
  social security administration forecasts,'' {\em Journal of Economic
  Perspectives}, vol.~29, pp.~239--58, May 2015.

\bibitem{JANSSEN2016371}
M.~Janssen and G.~Kuk, ``The challenges and limits of big data algorithms in
  technocratic governance,'' {\em Government Information Quarterly}, vol.~33,
  no.~3, pp.~371--377, 2016.
\newblock Open and Smart Governments: Strategies, Tools, and Experiences.

\bibitem{PSEs}
R.~Nabi, D.~Malinsky, and I.~Shpitser, ``Learning optimal fair policies,'' in
  {\em Proceedings of the 36th International Conference on Machine Learning}
  (K.~Chaudhuri and R.~Salakhutdinov, eds.), vol.~97 of {\em Proceedings of
  Machine Learning Research}, pp.~4674--4682, PMLR, 09--15 Jun 2019.

\bibitem{Disparate_impact_remover}
M.~Feldman, S.~A. Friedler, J.~Moeller, C.~Scheidegger, and
  S.~Venkatasubramanian, ``Certifying and removing disparate impact,'' in {\em
  Proceedings of the 21th ACM SIGKDD International Conference on Knowledge
  Discovery and Data Mining}, KDD '15, (New York, NY, USA), p.~259–268,
  Association for Computing Machinery, 2015.

\bibitem{adversarial_debiasing}
B.~H. Zhang, B.~Lemoine, and M.~Mitchell, ``Mitigating unwanted biases with
  adversarial learning,'' in {\em Proceedings of the 2018 AAAI/ACM Conference
  on AI, Ethics, and Society}, AIES '18, (New York, NY, USA), p.~335–340,
  Association for Computing Machinery, 2018.

\bibitem{balancing_competing_objectives}
E.~Rolf, M.~Simchowitz, S.~Dean, L.~T. Liu, D.~Bjorkegren, M.~Hardt, and
  J.~Blumenstock, ``Balancing competing objectives with noisy data: Score-based
  classifiers for welfare-aware machine learning,'' in {\em Proceedings of the
  37th International Conference on Machine Learning} (H.~D. III and A.~Singh,
  eds.), vol.~119 of {\em Proceedings of Machine Learning Research},
  pp.~8158--8168, PMLR, 13--18 Jul 2020.

\bibitem{gender_shades}
J.~Buolamwini and T.~Gebru, ``Gender shades: Intersectional accuracy
  disparities in commercial gender classification,'' in {\em Proceedings of the
  1st Conference on Fairness, Accountability and Transparency} (S.~A. Friedler
  and C.~Wilson, eds.), vol.~81 of {\em Proceedings of Machine Learning
  Research}, (New York, NY, USA), pp.~77--91, PMLR, 23--24 Feb 2018.

\bibitem{actionable_gender}
I.~D. Raji and J.~Buolamwini, ``Actionable auditing: Investigating the impact
  of publicly naming biased performance results of commercial ai products,'' in
  {\em Proceedings of the 2019 AAAI/ACM Conference on AI, Ethics, and Society},
  AIES '19, (New York, NY, USA), p.~429–435, Association for Computing
  Machinery, 2019.

\bibitem{machine_translation}
M.~O.~R. Prates, P.~H. Avelar, and L.~C. Lamb, ``Assessing gender bias in
  machine translation: a case study with {Google} {Translate},'' {\em Neural
  Computing and Applications}, vol.~32, pp.~6363--6381, May 2020.

\bibitem{bolukbasi_man_is_to_computer_programmer}
T.~Bolukbasi, K.-W. Chang, J.~Zou, V.~Saligrama, and A.~Kalai, ``Man is to
  computer programmer as woman is to homemaker? debiasing word embeddings,'' in
  {\em Proceedings of the 30th International Conference on Neural Information
  Processing Systems}, NIPS'16, (Red Hook, NY, USA), p.~4356–4364, Curran
  Associates Inc., 2016.

\bibitem{datta2015automated}
A.~Datta, M.~C. Tschantz, and A.~Datta, ``Automated experiments on ad privacy
  settings: A tale of opacity, choice, and discrimination,'' {\em Proceedings
  on privacy enhancing technologies}, vol.~2015, no.~1, pp.~92--112, 2015.

\bibitem{article}
D.~Cirillo, S.~Catuara~Solarz, C.~Morey, E.~Guney, L.~Subirats, S.~Mellino,
  A.~Gigante, A.~Valencia, M.~Rementeria, A.~Chadha, and N.~Mavridis, ``Sex and
  gender differences and biases in artificial intelligence for biomedicine and
  healthcare,'' {\em npj Digital Medicine}, vol.~3, 12 2020.

\bibitem{fair_housing_act_1968}
90th U.S.~Congress, {\em Fair Housing Act}.
\newblock 1968.

\bibitem{equal_credit_act}
93rd U.S.~Congress, {\em Equal Credit Opportunity Act}.
\newblock 1974.

\bibitem{verma_definitions_explained}
S.~Verma and J.~Rubin, ``Fairness definitions explained,'' in {\em Proceedings
  of the International Workshop on Software Fairness}, FairWare '18, (New York,
  NY, USA), p.~1–7, Association for Computing Machinery, 2018.

\bibitem{Gajane2017OnFF}
P.~Gajane, ``On formalizing fairness in prediction with machine learning,''
  {\em ArXiv}, vol.~abs/1710.03184, 2017.

\bibitem{ninareh}
N.~Mehrabi, F.~Morstatter, N.~Saxena, K.~Lerman, and A.~Galstyan, ``A survey on
  bias and fairness in machine learning,'' {\em CoRR}, vol.~abs/1908.09635,
  2019.

\bibitem{binns2019apparent}
R.~Binns, ``On the apparent conflict between individual and group fairness,''
  2019.

\bibitem{kusner2018counterfactual}
M.~J. Kusner, J.~R. Loftus, C.~Russell, and R.~Silva, ``Counterfactual
  fairness,'' 2018.

\bibitem{hardt2016equality}
M.~Hardt, E.~Price, and N.~Srebro, ``Equality of opportunity in supervised
  learning,'' 2016.

\bibitem{aif360-oct-2018}
R.~K.~E. Bellamy, K.~Dey, M.~Hind, S.~C. Hoffman, S.~Houde, K.~Kannan,
  P.~Lohia, J.~Martino, S.~Mehta, A.~Mojsilovic, S.~Nagar, K.~N. Ramamurthy,
  J.~Richards, D.~Saha, P.~Sattigeri, M.~Singh, K.~R. Varshney, and Y.~Zhang,
  ``{AI Fairness} 360: An extensible toolkit for detecting, understanding, and
  mitigating unwanted algorithmic bias,'' Oct. 2018.

\bibitem{huffpost}
J.~Chayes, ``How machine learning advances will improve the fairness of
  algorithms,'' {\em HuffPost}, 2017.

\bibitem{Dua:2019}
D.~Dua and C.~Graff, ``{UCI} machine learning repository,'' 2017.

\bibitem{mehrabi2019survey}
N.~Mehrabi, F.~Morstatter, N.~Saxena, K.~Lerman, and A.~Galstyan, ``A survey on
  bias and fairness in machine learning,'' {\em arXiv preprint
  arXiv:1908.09635}, 2019.

\bibitem{besse2020survey}
P.~Besse, E.~del Barrio, P.~Gordaliza, J.-M. Loubes, and L.~Risser, ``A survey
  of bias in machine learning through the prism of statistical parity for the
  adult data set,'' 2020.

\bibitem{kamishima_fairness-aware_2012}
T.~Kamishima, S.~Akaho, H.~Asoh, and J.~Sakuma, ``Fairness-{Aware} {Classifier}
  with {Prejudice} {Remover} {Regularizer},'' in {\em Machine {Learning} and
  {Knowledge} {Discovery} in {Databases}} (P.~A. Flach, T.~De~Bie, and
  N.~Cristianini, eds.), (Berlin, Heidelberg), pp.~35--50, Springer Berlin
  Heidelberg, 2012.

\bibitem{nabi2018fair}
R.~Nabi and I.~Shpitser, ``Fair inference on outcomes,'' 2018.

\bibitem{DBLP:journals/corr/ZhangWW16a}
L.~Zhang, Y.~Wu, and X.~Wu, ``A causal framework for discovering and removing
  direct and indirect discrimination,'' {\em CoRR}, vol.~abs/1611.07509, 2016.

\bibitem{NIPS2017_9a49a25d}
F.~Calmon, D.~Wei, B.~Vinzamuri, K.~Natesan~Ramamurthy, and K.~R. Varshney,
  ``Optimized pre-processing for discrimination prevention,'' in {\em Advances
  in Neural Information Processing Systems} (I.~Guyon, U.~V. Luxburg,
  S.~Bengio, H.~Wallach, R.~Fergus, S.~Vishwanathan, and R.~Garnett, eds.),
  vol.~30, Curran Associates, Inc., 2017.

\bibitem{zhang}
B.~H. Zhang, B.~Lemoine, and M.~Mitchell, ``Mitigating unwanted biases with
  adversarial learning,'' in {\em Proceedings of the 2018 AAAI/ACM Conference
  on AI, Ethics, and Society}, AIES '18, (New York, NY, USA), p.~335–340,
  Association for Computing Machinery, 2018.

\bibitem{Zafar}
M.~B. Zafar, I.~Valera, M.~G. Rogriguez, and K.~P. Gummadi, ``{Fairness
  Constraints: Mechanisms for Fair Classification},'' in {\em Proceedings of
  the 20th International Conference on Artificial Intelligence and Statistics}
  (A.~Singh and J.~Zhu, eds.), vol.~54 of {\em Proceedings of Machine Learning
  Research}, (Fort Lauderdale, FL, USA), pp.~962--970, PMLR, 20--22 Apr 2017.

\bibitem{wu2018fairnessaware}
Y.~Wu, L.~Zhang, and X.~Wu, ``Fairness-aware classification: Criterion,
  convexity, and bounds,'' 2018.

\bibitem{post_eq_calib}
G.~Pleiss, M.~Raghavan, F.~Wu, J.~Kleinberg, and K.~Q. Weinberger, ``On
  fairness and calibration,'' in {\em Advances in Neural Information Processing
  Systems} (I.~Guyon, U.~V. Luxburg, S.~Bengio, H.~Wallach, R.~Fergus,
  S.~Vishwanathan, and R.~Garnett, eds.), vol.~30, Curran Associates, Inc.,
  2017.

\bibitem{compas_paper}
S.~Corbett-Davies, E.~Pierson, A.~Feller, S.~Goel, and A.~Huq, ``Algorithmic
  decision making and the cost of fairness,'' in {\em Proceedings of the 23rd
  ACM SIGKDD International Conference on Knowledge Discovery and Data Mining},
  KDD '17, (New York, NY, USA), p.~797–806, Association for Computing
  Machinery, 2017.

\bibitem{propublica_story}
J.~Angwin, J.~Larson, S.~Mattu, and L.~Kirchner, ``Machine bias: There’s
  software used across the country to predict future criminals. and it’s
  biased against blacks.,'' {\em ProPublica}, 2016.

\bibitem{DBLP:journals/corr/abs-1807-00199}
C.~Wadsworth, F.~Vera, and C.~Piech, ``Achieving fairness through adversarial
  learning: an application to recidivism prediction,'' {\em CoRR},
  vol.~abs/1807.00199, 2018.

\bibitem{pmlr-v80-komiyama18a}
J.~Komiyama, A.~Takeda, J.~Honda, and H.~Shimao, ``Nonconvex optimization for
  regression with fairness constraints,'' in {\em Proceedings of the 35th
  International Conference on Machine Learning} (J.~Dy and A.~Krause, eds.),
  vol.~80 of {\em Proceedings of Machine Learning Research},
  (Stockholmsmässan, Stockholm Sweden), pp.~2737--2746, PMLR, 10--15 Jul 2018.

\bibitem{DBLP:journals/corr/abs-1903-08136}
N.~Mehrabi, F.~Morstatter, N.~Peng, and A.~Galstyan, ``Debiasing community
  detection: The importance of lowly-connected nodes,'' {\em CoRR},
  vol.~abs/1903.08136, 2019.

\bibitem{pre-process}
F.~Kamiran and T.~Calders, ``Data preprocessing techniques for classification
  without discrimination,'' {\em Knowl. Inf. Syst.}, vol.~33, p.~1–33, Oct.
  2012.

\bibitem{lowd}
D.~Lowd and C.~Meek, ``Adversarial learning,'' in {\em Proceedings of the
  Eleventh ACM SIGKDD International Conference on Knowledge Discovery in Data
  Mining}, KDD '05, (New York, NY, USA), p.~641–647, Association for
  Computing Machinery, 2005.

\bibitem{Sweeney}
C.~Sweeney and M.~Najafian, ``Reducing sentiment polarity for demographic
  attributes in word embeddings using adversarial learning,'' in {\em
  Proceedings of the 2020 Conference on Fairness, Accountability, and
  Transparency}, FAT* '20, (New York, NY, USA), p.~359–368, Association for
  Computing Machinery, 2020.

\bibitem{FairnessGAN}
P.~{Sattigeri}, S.~C. {Hoffman}, V.~{Chenthamarakshan}, and K.~R. {Varshney},
  ``Fairness gan: Generating datasets with fairness properties using a
  generative adversarial network,'' {\em IBM Journal of Research and
  Development}, vol.~63, no.~4/5, pp.~3:1--3:9, 2019.

\bibitem{goodfellow2014generative}
I.~J. Goodfellow, J.~Pouget-Abadie, M.~Mirza, B.~Xu, D.~Warde-Farley, S.~Ozair,
  A.~Courville, and Y.~Bengio, ``Generative adversarial networks,'' 2014.

\bibitem{gencoglu}
O.~{Gencoglu}, ``Cyberbullying detection with fairness constraints,'' {\em IEEE
  Internet Computing}, vol.~25, no.~1, pp.~20--29, 2021.

\bibitem{DBLP:journals/corr/ZhaoWYOC17}
J.~Zhao, T.~Wang, M.~Yatskar, V.~Ordonez, and K.~Chang, ``Men also like
  shopping: Reducing gender bias amplification using corpus-level
  constraints,'' {\em CoRR}, vol.~abs/1707.09457, 2017.

\bibitem{woodworth2017learning}
B.~Woodworth, S.~Gunasekar, M.~I. Ohannessian, and N.~Srebro, ``Learning
  non-discriminatory predictors,'' 2017.

\bibitem{decision_theory}
F.~Kamiran, A.~Karim, and X.~Zhang, ``Decision theory for discrimination-aware
  classification,'' in {\em 2012 IEEE 12th International Conference on Data
  Mining}, pp.~924--929, 2012.

\bibitem{lohia2018bias}
P.~K. Lohia, K.~N. Ramamurthy, M.~Bhide, D.~Saha, K.~R. Varshney, and R.~Puri,
  ``Bias mitigation post-processing for individual and group fairness,'' 2018.

\bibitem{neural_popular}
N.~M. Nawi, W.~H. Atomi, and M.~Rehman, ``The effect of data pre-processing on
  optimized training of artificial neural networks,'' {\em Procedia
  Technology}, vol.~11, pp.~32--39, 2013.
\newblock 4th International Conference on Electrical Engineering and
  Informatics, ICEEI 2013.

\bibitem{lucas2021analyzing}
J.~Lucas, J.~Bae, M.~R. Zhang, S.~Fort, R.~Zemel, and R.~Grosse, ``Analyzing
  monotonic linear interpolation in neural network loss landscapes,'' 2021.

\bibitem{bird2020fairlearn}
S.~Bird, M.~Dudík, R.~Edgar, B.~Horn, R.~Lutz, V.~Milan, M.~Sameki,
  H.~Wallach, and K.~Walker, ``Fairlearn: A toolkit for assessing and improving
  fairness in ai,'' Tech. Rep. MSR-TR-2020-32, Microsoft, May 2020.

\bibitem{ahn2019fairsight}
Y.~Ahn and Y.-R. Lin, ``Fairsight: Visual analytics for fairness in decision
  making,'' {\em IEEE transactions on visualization and computer graphics},
  vol.~26, no.~1, pp.~1086--1095, 2019.

\bibitem{2018aequitas}
P.~Saleiro, B.~Kuester, A.~Stevens, A.~Anisfeld, L.~Hinkson, J.~London, and
  R.~Ghani, ``Aequitas: A bias and fairness audit toolkit,'' {\em arXiv
  preprint arXiv:1811.05577}, 2018.

\bibitem{rbg}
BBC, ``Ruth {Bader} {Ginsburg} in pictures and her own words,'' {\em BBC News},
  Sept. 2020.

\bibitem{diversity_pedagogy}
V.~Pournaghshband and P.~Medel, ``Promoting diversity-inclusive computer
  science pedagogies: A multidimensional perspective,'' in {\em Proceedings of
  the 2020 ACM Conference on Innovation and Technology in Computer Science
  Education}, ITiCSE '20, (New York, NY, USA), p.~219–224, Association for
  Computing Machinery, 2020.

\end{thebibliography}
\end{document}